\documentclass[sigconf]{acmart}
\AtBeginDocument{%
  }

\copyrightyear{2026}
\acmYear{2026}
\setcopyright{cc}
\setcctype{by}
\acmConference[MM '26]{Proceedings of the 34th ACM International Conference on Multimedia}{November 10--14, 2026}{Rio de Janeiro, Brazil}
\acmBooktitle{Proceedings of the 34th ACM International Conference on Multimedia (MM '26), November 10--14, 2026, Rio de Janeiro, Brazil}
\acmDOI{10.1145/3767308.3835061}
\acmISBN{979-8-4007-2213-4/2026/11}

\usepackage{subcaption}  
\usepackage{multirow}   
\usepackage{makecell}   
\definecolor{pubcolor}{HTML}{e67e22}
\newcommand{\pub}[1]{\textcolor{pubcolor}{\scriptsize{~[#1]}}}

\begin{document}

\title{MuRA: Multi-Rank Adaptation for Efficient and Effective Test-Time Vision-Language Generalization}

\author{Gengyuan Liu}
\authornote{These authors contributed equally to this research.}
\orcid{0009-0002-7008-4281}
\affiliation{%
  \institution{Tsinghua University}
  \city{Shenzhen}
  \country{China}
}
\email{gy-liu24@mails.tsinghua.edu.cn}

\author{Nanzhou Wang}
\authornotemark[1]
\orcid{0009-0004-8878-3624}
\affiliation{%
  \institution{Tsinghua University}
  \city{Shenzhen}
  \country{China}
}
\email{wnz24@mails.tsinghua.edu.cn}

\author{Chang Liu}
\authornote{Corresponding authors.}
\orcid{0000-0001-6747-0646}
\affiliation{%
  \institution{Tsinghua University}
  \city{Beijing}
  \country{China}
}
\email{liuchang2022@tsinghua.edu.cn}

\author{Qinwen Wu}
\orcid{0009-0008-8613-8137}
\affiliation{%
  \institution{Tsinghua University}
  \city{Shenzhen}
  \country{China}
}
\email{wuqw24@mails.tsinghua.edu.cn}

\author{Zhenhao Wang}
\orcid{0009-0002-7997-9671}
\affiliation{%
  \institution{Huazhong University of Science and Technology}
  \city{Wuhan}
  \country{China}
}
\email{m202470074@hust.edu.cn}

\author{Jiacong Wang}
\orcid{0009-0001-8719-0614}
\affiliation{%
  \institution{University of Chinese Academy of Sciences}
  \city{Beijing}
  \country{China}
}
\email{wangjiacong20@mails.ucas.ac.cn}

\author{Bokui Chen}
\orcid{0000-0002-4947-5619}
\authornotemark[2]
\affiliation{%
  \institution{Tsinghua University}
  \city{Shenzhen}
  \country{China}
}
\email{chenbk@tsinghua.edu.cn}

\author{Xiangyang Ji}
\authornotemark[2]
\orcid{0000-0001-9542-5260}
\affiliation{%
  \institution{Tsinghua University}
  \city{Beijing}
  \country{China}
}
\email{xyji@tsinghua.edu.cn}

\renewcommand{\shortauthors}{Gengyuan Liu et al.}

\begin{abstract}
  Vision-language models exhibit remarkable zero-shot capabilities but suffer significant performance degradation under distribution shifts. While test-time adaptation (TTA) via Low-Rank Adaptation offers a parameter-efficient solution, we identify a fundamental bottleneck in current methods: the reliance on static rank configurations. Because visual inputs inherently possess varying information densities, a fixed rank forces an inevitable optimization compromise, leading to underfitting on complex scenes and overfitting on simple ones. To bridge this gap, we propose Multi-Rank Adaptation (MuRA), a novel framework that dynamically selects and fuses adaptation modules of varying capacities based on token-level visual complexity. MuRA synergizes Multi-Rank Orthogonal Decomposition to provide a superior, knowledge-preserving initialization, and Unified Component Fusion with Continuous Router Updating to sustainably learn semantic-to-rank mappings. Furthermore, we provide rigorous theoretical justifications mathematically proving the necessity and gradient stability of this adaptive mechanism. Crucially, MuRA's dynamic design uniquely thrives at the deepest visual layer, capitalizing on the shortest gradient backpropagation path. Extensive experiments demonstrate that MuRA achieves state-of-the-art accuracy across extensive domain generalization and cross-dataset benchmarks while significantly reducing both computational and memory overhead.
\end{abstract}







\maketitle

\section{Introduction}

\begin{figure*}[t]
    \centering
    \includegraphics[width=1\linewidth]{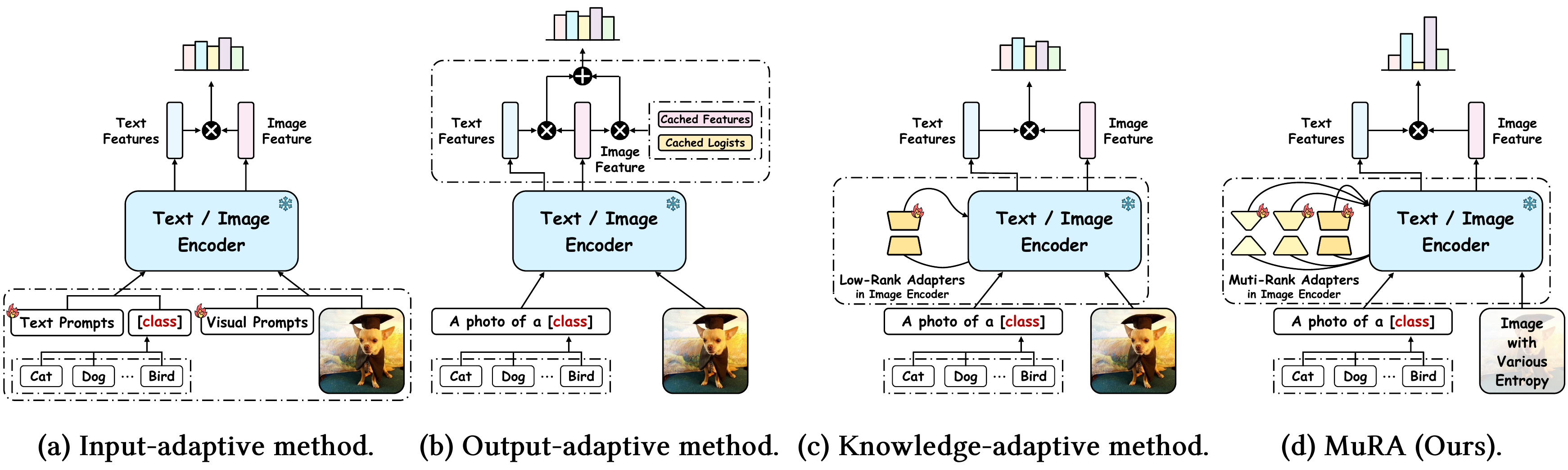}
    \caption{\textbf{Overview of test-time adaptation approaches in vision-language models.}}
    \label{fig:ttl_comparison}
\end{figure*}

\begin{figure}[t] 
    \centering
    
    \begin{minipage}{0.23\textwidth} 
        \centering
        \includegraphics[width=\linewidth]{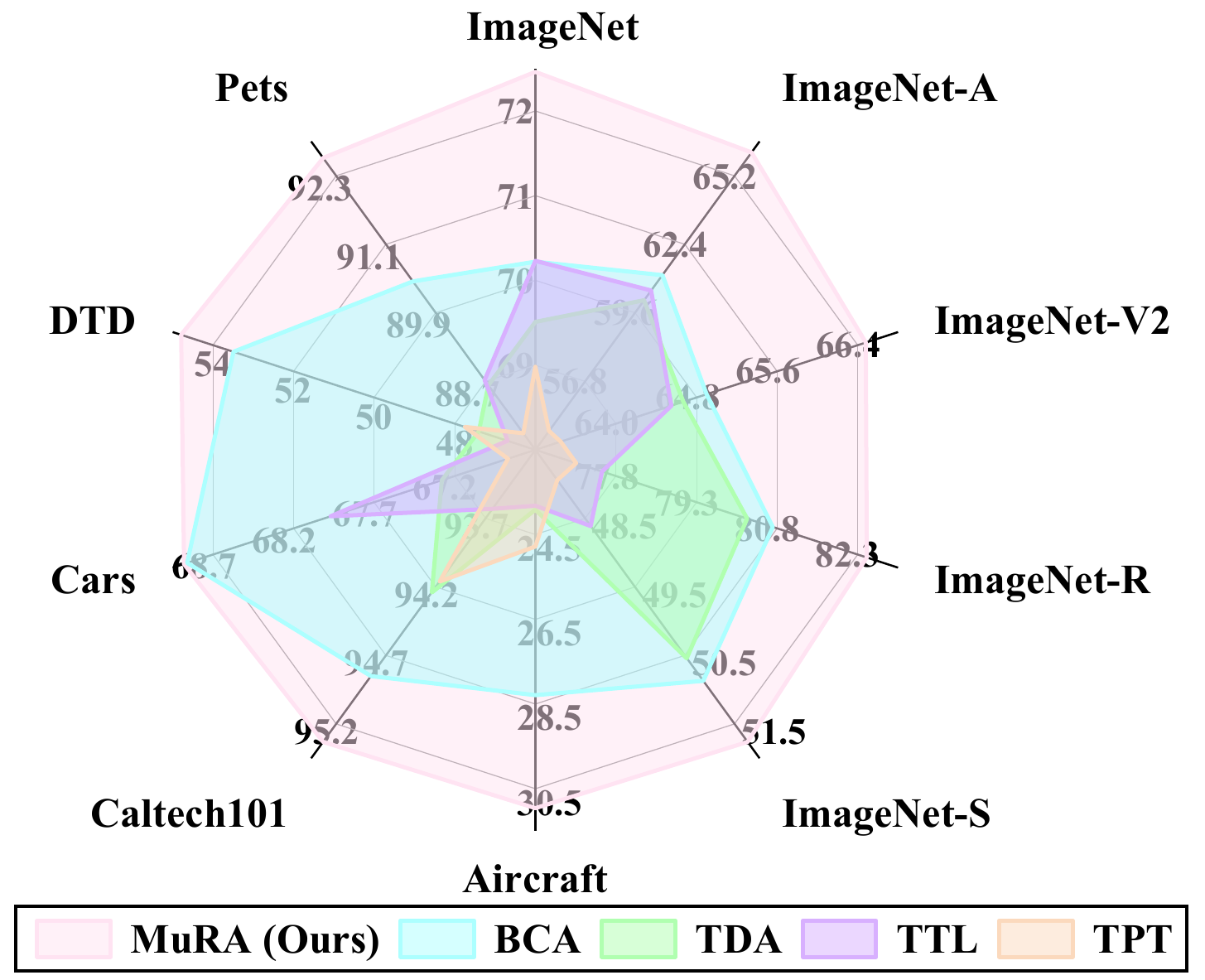}
    \end{minipage}
    \hfill 
    \begin{minipage}{0.23\textwidth}
        \centering
        \includegraphics[width=\linewidth]{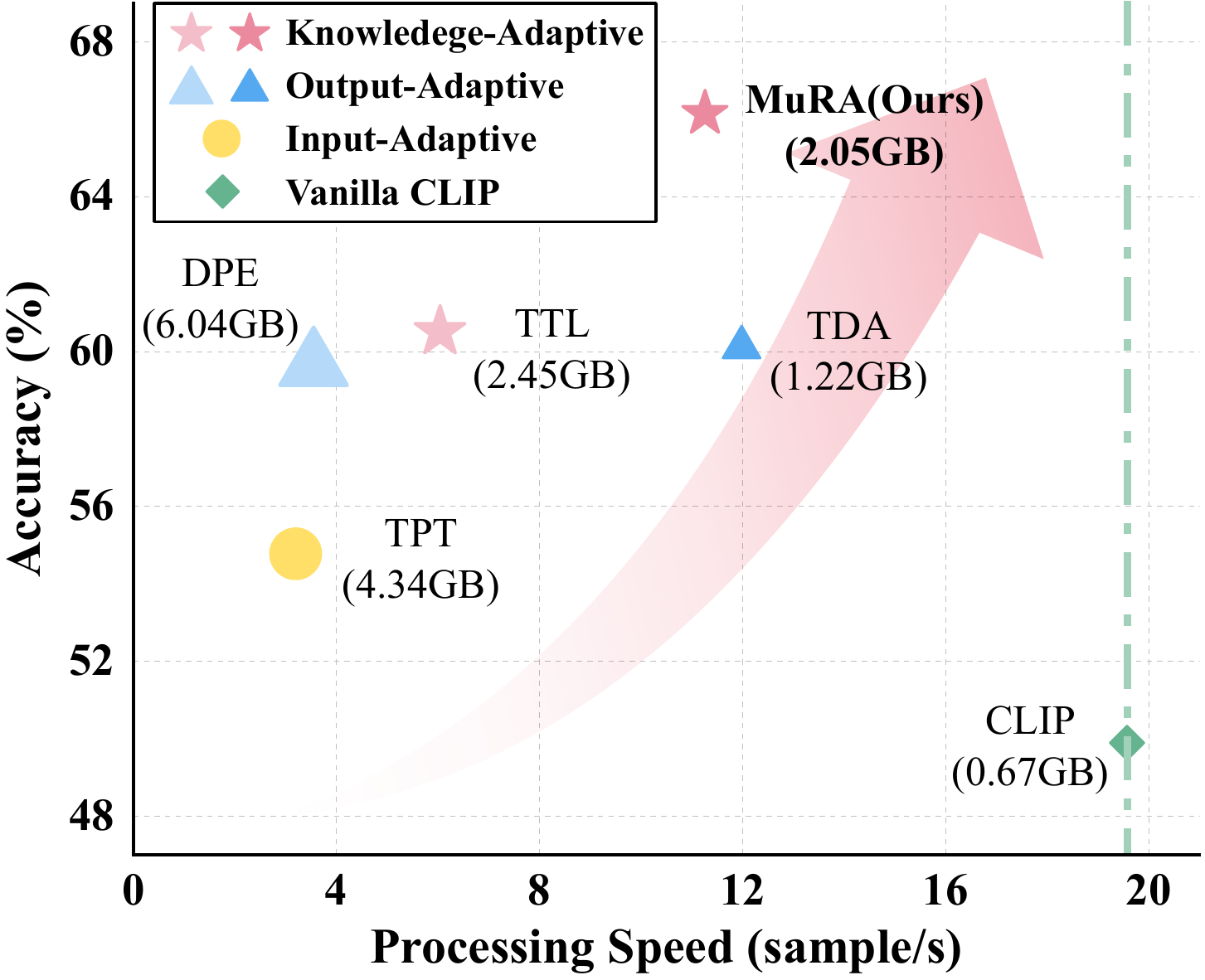}
    \end{minipage}
    
    \caption{
         \textbf{Effectiveness and efficiency of MuRA.} 
        (\textbf{Left}) MuRA consistently surpasses existing methods across diverse datasets. 
        (\textbf{Right}) MuRA delivers higher throughput and a reduced memory footprint compared to competing baselines.}
    \label{fig:eff}
\end{figure}

Large-scale vision-language models (VLMs) such as CLIP~\citep{radford2021learning} achieve strong zero-shot visual recognition by training on massive image-text pairs and learning aligned representations in a shared embedding space. However, their performance degrades severely under distribution shifts, a vulnerability that Test-Time Adaptation (TTA) mitigates during inference using exclusively unlabeled test samples, bypassing the need for costly target annotations.

As illustrated in Figure~\ref{fig:ttl_comparison}, existing TTA methods for VLMs generally fall into three categories: input-adaptive, output-adaptive, and knowledge-adaptive approaches. Input-adaptive methods optimize learnable prompts or transformations at test time, while output-adaptive approaches recalibrate the final predictions using cached feature statistics. More recently, knowledge-adaptive methods have sought deeper model adaptation by injecting lightweight learnable modules, such as Low-Rank Adaptation (LoRA), optimizing them directly during inference. While LoRA-based TTA offers a promising pathway by explicitly updating visual representations, current implementations yield limited performance gains and can even trigger catastrophic degradation under severe distribution shifts.

\begin{figure}[t] 
    \centering
    
    \begin{minipage}{0.235\textwidth}
        \centering
        \includegraphics[width=\linewidth]{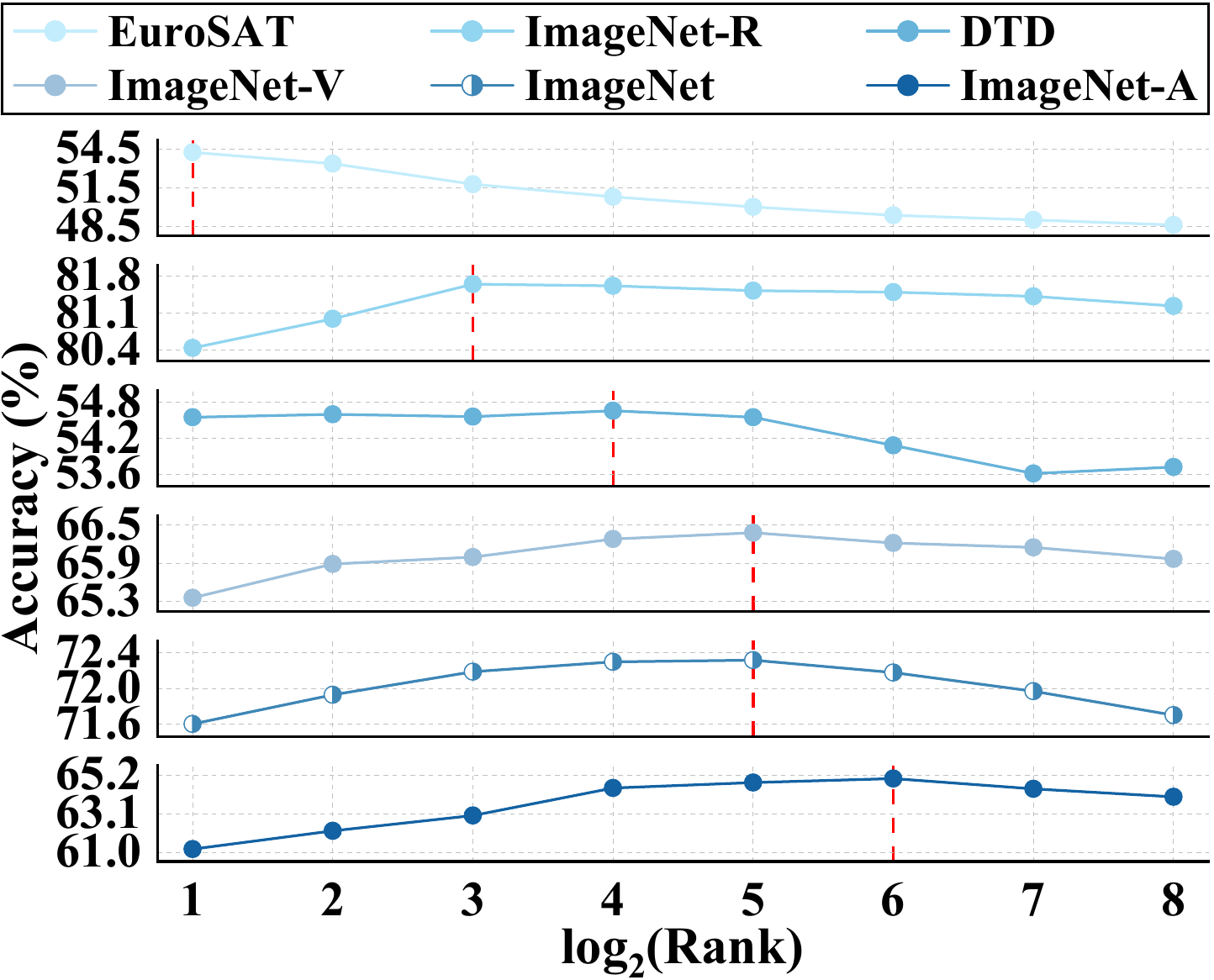}
    \end{minipage}
    \hfill 
    \begin{minipage}{0.235\textwidth}
        \centering
        \includegraphics[width=\linewidth]{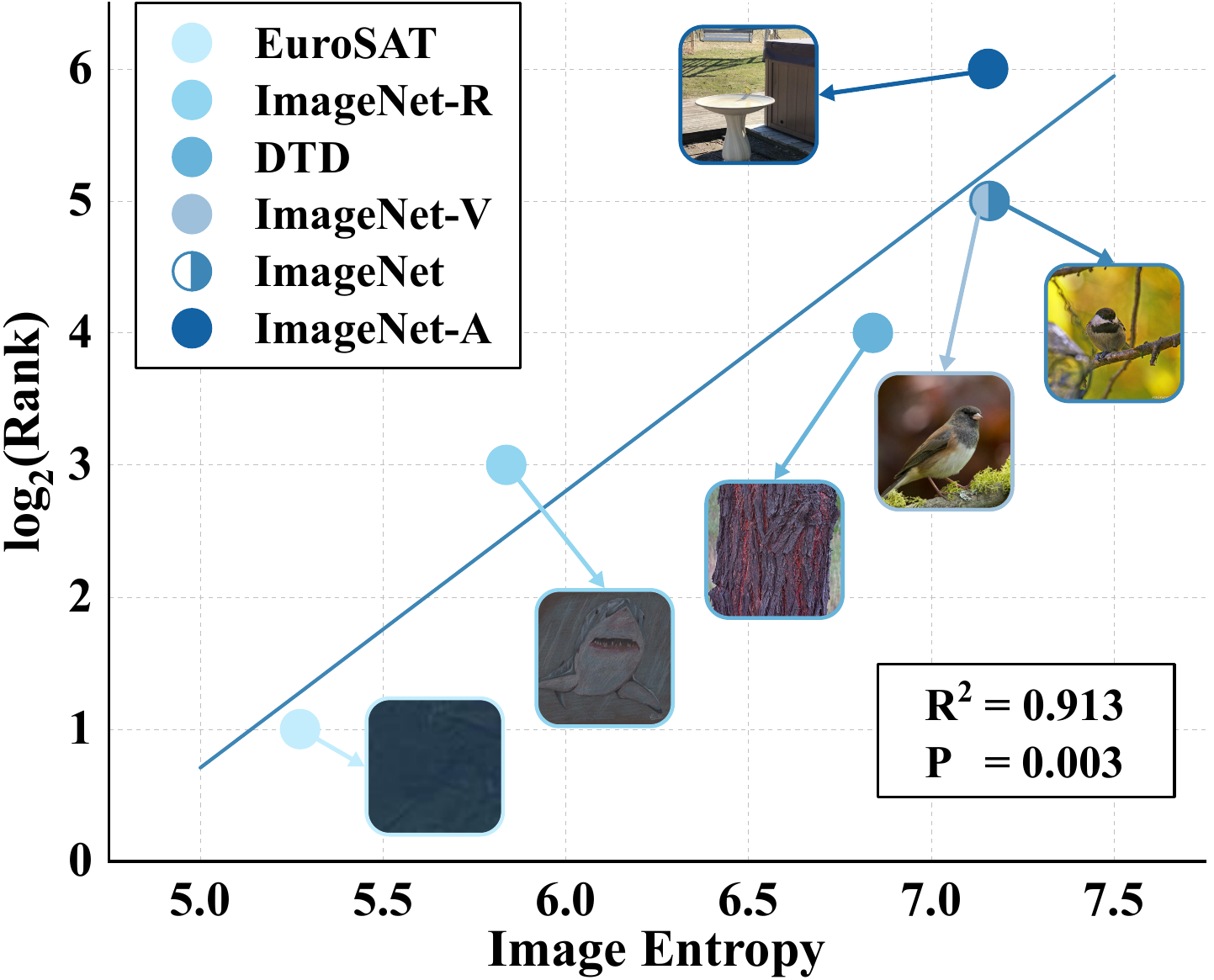}
    \end{minipage}
    
    \caption{Motivation for dynamic rank adaptation. (Left) The optimal static rank varies significantly across six distinct datasets. (Right) A strong positive correlation ($R^2 = 0.913$) exists between this optimal rank ($\log_2$) and data complexity.}
    \label{fig:motivation} 
\end{figure}

Through systematic evaluation, we identify the fundamental bottleneck of current knowledge-adaptive TTA: the static rank configuration. Following the standard parameter-efficient fine-tuning paradigm, existing methods universally assign a fixed rank to all test samples. However, our analysis reveals that the optimal representational capacity required for adaptation varies drastically across different data distributions. As illustrated in Figure~\ref{fig:motivation} (Left), distinct domains exhibit entirely different rank preferences. Furthermore, we discover a strong linear correlation ($R^2 = 0.913$) between the visual complexity of an input—quantified by image entropy~\citep{wu2013local} across RGB channels—and its optimal LoRA rank (Figure~\ref{fig:motivation}, Right). Low-entropy inputs, such as simple textures or stylized art, require only low-rank updates; high-entropy inputs, such as visually cluttered or corrupted scenes, demand higher ranks to capture complex structural semantics. A static rank forces an inevitable optimization compromise, leading to underfitting on complex scenes and overfitting on simple ones. While dynamic rank selection has been explored in offline fine-tuning, these methods rely on prolonged training with labeled data, rendering them entirely inapplicable to the label-free, single-pass constraints of TTA.

To bridge this critical gap, we propose Multi-Rank Adaptation (MuRA), an efficient and effective TTA framework that dynamically selects and fuses LoRA modules of varying ranks based on token-level visual complexity. MuRA introduces two core innovations. First, Multi-Rank Orthogonal Decomposition (MROD) initializes rank-specific adaptation modules by decomposing pretrained weights into principal components. This directly solves the optimization instability and vanishing gradients inherent in standard zero-initialized LoRA during short-horizon test-time updates. Second, Unified Component Fusion (UCF) employs a lightweight, learnable router to dynamically compute token-level mixing weights. Unlike conventional TTA methods that aggressively reset parameters after every sample, we introduce a Continuous Router Updating (CRU) strategy. By explicitly resetting only the capacity anchors while continuously updating the router, CRU allows the model to globally learn the generalized "complexity-to-capacity" mapping.

Crucially, MuRA resolves the effectiveness-efficiency dilemma inherent in conventional methods. While standard single-rank adaptation suffers severe performance drops when restricted to the deepest visual layer, MuRA's dynamic routing naturally excels at encoding these complex, high-level semantics. By uniquely achieving its peak accuracy solely at the deepest layer, MuRA capitalizes on the shortest gradient backpropagation path. As shown in Figure~\ref{fig:eff}, this structural synergy enables MuRA to deliver superior accuracy with higher throughput and a significantly smaller memory footprint than competing baselines.

Our main contributions are summarized as follows:
\begin{itemize}
    \item We identify the fundamental limitation of static rank configurations in knowledge-adaptive TTA, empirically establishing a strong correlation between input visual complexity (entropy) and the optimal adaptation rank.
    \item We propose MuRA, a novel TTA framework that synergizes Multi-Rank Orthogonal Decomposition, Unified Component Fusion, and Continuous Router Updating for dynamic rank routing. We further provide rigorous theoretical justifications proving the necessity and gradient stability of MuRA.
    \item Extensive experiments demonstrate that MuRA's dynamic design uniquely thrives at the deepest visual layer, achieving state-of-the-art accuracy across multiple generalization benchmarks while fundamentally minimizing computational overhead and memory footprint.
\end{itemize}


\section{Related Work}

\subsection{Vision-Language Model Adaptation}
With the rapid surge of VLMs \cite{radford2021learning, li2022blip, liu2023visual, wang2024qwen2, wang2025vgr, wang2024world, lei2025scalability}, adapting these large-scale pre-trained models to downstream tasks without full fine-tuning has become crucial for efficient deployment. Adaptation methods of VLMs mainly fall into prompt tuning-based and feature adapter-based ones. Prompt tuning has emerged as a prominent strategy, particularly effective in few-shot and test-time scenarios. For example, CoOp~\citep{zhou2022learning} optimizes learnable continuous prompts appended to class names, while CoCoOp~\citep{zhou2022conditional} introduces a meta-network that generates image-conditioned prompts based on visual features. UPT~\citep{zang2022unified} jointly learns both visual and textual prompts through a lightweight network. For feature adapter, it modifies the internal representation of VLMs by inserting lightweight trainable modules. CLIP-Adapter~\citep{gao2024clip} adds a residual bottleneck layer to fuse task-specific features, and Tip-Adapter~\citep{zhang2021tip} uses a non-parametric cache for similarity-based classification. These methods are parameter-efficient and well-suited to low-data scenarios.

\subsection{Test-time adaptation with vision-language models}

Traditional TTA approaches in computer vision have explored various unsupervised techniques, including entropy minimization, batch normalization calibration, pseudo-labeling, and consistency regularization.
In the context of VLMs, several adaptation strategies have emerged. Prompt-based methods like TPT~\citep{shu2022test} optimize text prompts by minimizing marginal entropy across augmented views. DiffTPT~\citep{feng2023diverse} enhances prompt robustness using diffusion-based augmentations, while PromptAlign~\citep{abdul2023align} explicitly aligns token distributions between test and source domains.
Beyond textual prompts, LoRA-based adaptation has gained traction for its efficiency, with TTL~\citep{imam2025test} adapting CLIP's attention weights during inference.
In contrast, training-free methods avoid backpropagation entirely: TDA~\citep{karmanov2024efficient} refines pseudo-labels via a key-value cache, MTA~\citep{zanella2024test} filters reliable augmentations through inlierness optimization, and DPE~\citep{zhang2024dual} maintains dual semantic and visual prototypes.
Recent efficient methods include MCP~\citep{chen2025multi} using multi-cache for intra-class compactness, TT-RAA~\citep{fan2025test} leveraging streaming Gaussian databases for retrieval, and GS-Bias~\citep{huang2025gs} learning global-spatial biases for logit calibration.
BCA~\citep{zhou2025bayesian} introduces an adaptive prior mechanism that updates class embeddings using sample posteriors.

\section{Method}
\label{sec:method}
\subsection{Overview}
\begin{figure*}[t]
    \centering
    \includegraphics[width=1\linewidth]{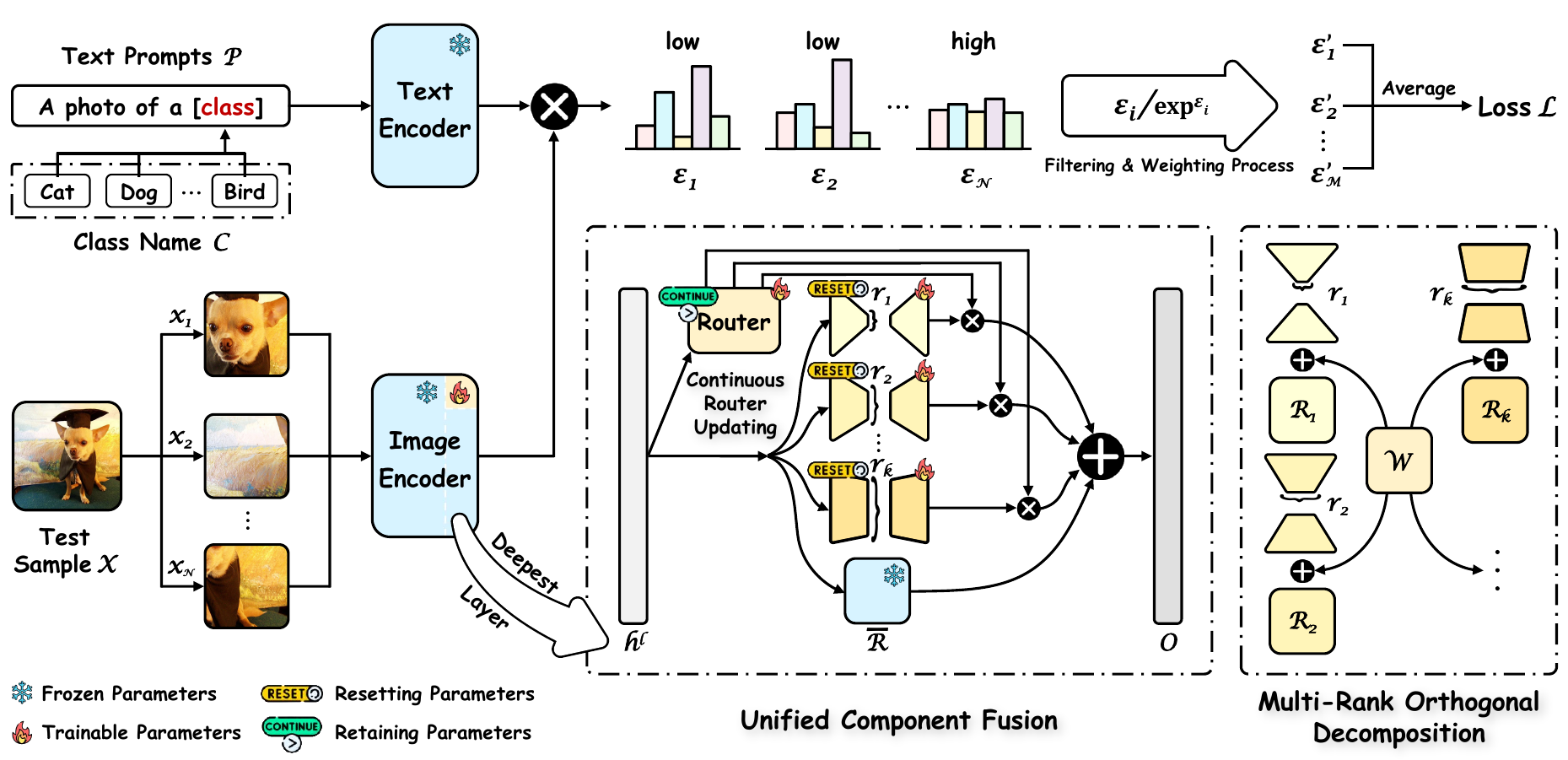}
    \caption{
        \textbf{Overall Architecture of Multi-Rank Adaptation (MuRA).} 
        MuRA is an adaptive framework integrated into the deepest visual layer, designed to overcome the limitations of fixed-rank adaptation. It is built upon two core innovation components: 
        (\textbf{1}) \textbf{Multi-Rank Orthogonal Decomposition (MROD)}, which initializes rank-specific adaptation components $\{A_i, B_i\}$ and residual matrices $\{R_i\}$ from the pre-trained weight $W$.
        (\textbf{2}) \textbf{Unified Component Fusion (UCF)}, which employs a continuously updated Router (CRU) to dynamically combine these different rank components for token-level adaptation. 
        The overall process is optimized via an entropy-based test-time adaptation objective applied to augmented views.
    }
    \label{fig:method}
\end{figure*}

As shown in Figure~\ref{fig:method}, MuRA is an adaptive framework integrated at the deepest visual layer that learns dynamic rank preferences across diverse distributions. Multi-Rank Orthogonal Decomposition (MROD) orthogonally decomposes pretrained weights to initialize different-rank LoRA modules, avoiding optimization difficulty from zero-initialization. Unified Component Fusion (UCF) averages residuals $\{R_i\}_{i=1}^k$ into $\bar{R}$, then uses a Mixture-of-Experts (MoE) mechanism to construct and combine different low-rank matrices. Continuous Router Updating (CRU) continuously updates the router across samples, capturing visual-semantic correlations for token-specific rank selection.

\begin{figure}[t]
    \centering
    \begin{subfigure}[t]{0.265\textwidth}
        \centering
        \includegraphics[width=\linewidth]{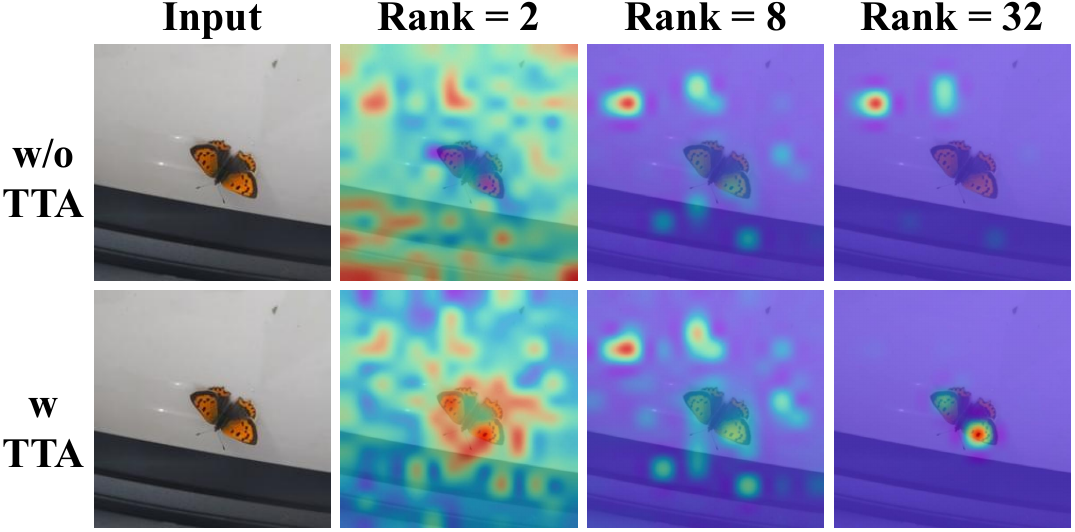}
        \caption{Attention visualization.}
        \label{fig:rank_pre}
    \end{subfigure}%
    \hfill
    \begin{subfigure}[t]{0.206\textwidth}
        \centering
        \includegraphics[width=\linewidth]{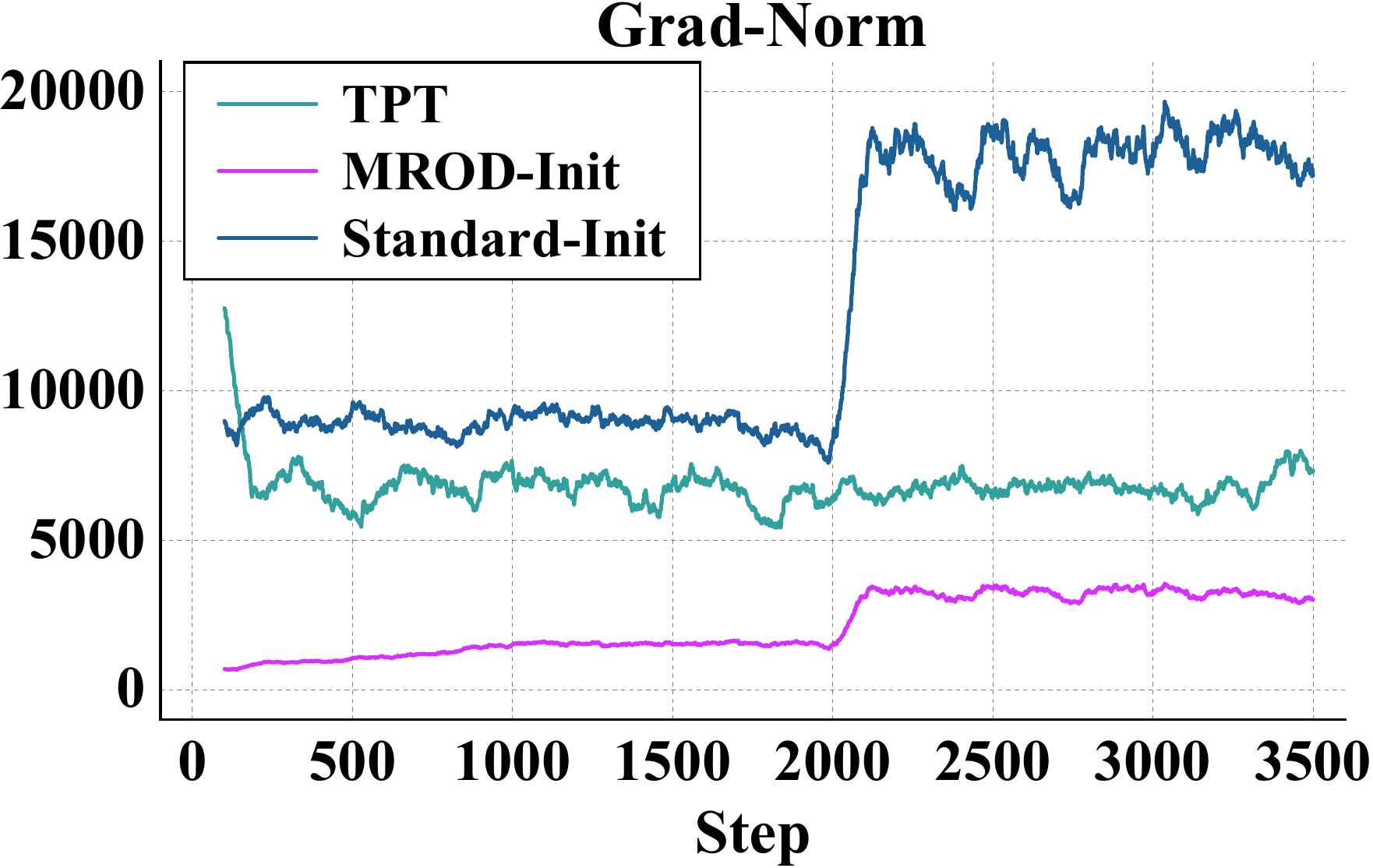}
        \caption{Gradient norm analysis.}
        \label{fig:grad}
    \end{subfigure}
    \caption{\textbf{The stabilizing effect of MROD-Initialization and Multi-Rank attention refinement.} (a) Visualizing LoRA attention across ranks. (b) Gradient Norm analysis during TTA.}
    \label{fig:combined}
\end{figure}

For test-time adaptation on a given input image $x$, we apply a data augmentation function $\mathcal{A}$ to generate a set of augmented views. For each view $\tilde{x} \in \mathcal{A}(x)$, we compute the prediction entropy and apply an adaptive weighting scheme to balance their contributions, following the approach in~\citep{imam2025test}. The trainable parameters are updated by minimizing the final loss computed as Eq.~\ref{eq:loss}:
\begin{equation}
\label{eq:entropy}
\mathcal{H}_{\Phi}(\tilde{x}) = - \sum_{j=1}^C \tilde{p}_{\Phi}(y_j \mid \tilde{x}) \log \tilde{p}_{\Phi}(y_j \mid \tilde{x}),
\end{equation}
\begin{equation}
\label{eq:weight}
\beta_{\Phi}(\tilde{x}) = \frac{1}{\exp(\mathcal{H}_{\Phi}(\tilde{x}) - \epsilon)},
\end{equation}
\begin{equation}
\label{eq:loss}
\mathcal{L} = - \frac{1}{\rho N} \sum_{\tilde{x} \in \mathcal{A}(X)} \mathbf{1}[\mathcal{H}_{\Phi}(\tilde{x}) \leq \tau] \cdot \beta_{\Phi}(\tilde{x}) \cdot \mathcal{H}_{\Phi}(\tilde{x}),
\end{equation}
where $C$ denotes the total number of classes, $\tilde{p}_{\Phi}$ denotes the predicted class probabilities, $\beta_{\Phi}(\tilde{x})$ represents the adaptive weighting term as defined in Eq.~\ref{eq:weight}, and $\tau$ is a dynamic threshold set as the $\rho$-percentile of the entropy values across all augmented views.

Upon completion of test-time adaptation, the VLM performs inference on the original image without augmentation. Subsequently, the trainable low-rank matrices are reset to their MROD-initialized values prepared for the next adaptation while preserving the router's accumulated knowledge.



\subsection{Main Components}
\paragraph{Multi-Rank Orthogonal Decomposition.}
Following~\citep{meng2024pissa}, we initialize LoRA modules with low-rank principal components from orthogonally decomposed pretrained weights. Figure~\ref{fig:rank_pre} shows that different ranks capture distinct, meaningful attention patterns before adaptation, providing effective knowledge representations; after TTA, these maps focus more strongly on salient image regions. Gradient analysis in Figure~\ref{fig:grad} further shows smaller, more stable norms than standard LoRA and input-adaptive TPT, reducing adaptation instability. Motivated by these observations, we propose MROD to initialize LoRA parameters through structured decomposition of pretrained weights. Given rank configurations \{$r_1,...,r_k$\} in ascending order, for a weight matrix $W \in \mathbb{R}^{m \times n}$, MROD performs its economic singular value decomposition: $W = USV^\top$, where $U \in \mathbb{R}^{m \times \min(m,n)}$, $V \in \mathbb{R}^{n \times \min(m,n)}$ are orthogonal matrices, and $S = \mathrm{diag}(\mathbf{s})$ contains singular values in descending order.
For each rank $r_i$, we initialize the LoRA projection matrices as follows:

\begin{equation}
A_i = U_{[:, :r_i]} S_{[:r_i, :r_i]}^{1/2} \in \mathbb{R}^{m \times r_i},
\end{equation}
\begin{equation}
B_i = S_{[:r_i, :r_i]}^{1/2} V_{[:, :r_i]}^\top \in \mathbb{R}^{r_i \times n},
\end{equation}
\begin{equation}
R_i = U_{[:, r_i:]} S_{[r_i:, r_i:]} V_{[:, r_i:]}^\top \in \mathbb{R}^{m \times n},
\end{equation}
where $R_i$ represents the residual matrix after decomposition. With these initialization strategies, we can decompose $W$ into:
\begin{equation}
W = A_i B_i^\top + R_i.
\end{equation}

This orthogonal decomposition ensures $A_iB_i \perp R_i$ for each $i$, facilitating more efficient optimization compared to zero-initialization. Note that MROD is performed only once, with these initialized projection matrices cached for subsequent resets, incurring negligible computational overhead.

\paragraph{Unified Component Fusion.}
To effectively combine different-rank LoRA modules and residual matrices, we propose UCF, drawing inspiration from Mixture-of-Experts~\citep{jacobs1991adaptive}.

The residual matrices from MROD are aggregated through averaging: $\bar{\mathit{R}} = \frac{1}{k} \sum_{i=1}^k \mathit{R}_i$. The aggregated matrix $\bar{R}$ replaces the original pretrained weight and remains frozen to maintain pretrained consistency. Let $\mathit{h}_l$ denote the token representation from the deepest layer. A zero-initialized router $\mathit{W}_r \in \mathbb{R}^{k \times d}$ computes rank preferences through: $\boldsymbol{\pi}(\mathit{h}_l) = \mathrm{softmax}(\mathit{W}_r \mathit{h}_l)$, where $d$ denotes the feature dimension. The router initialization ensures stable initialization by producing uniform softmax outputs.
The final feature output is $\mathit{o} = \sum_{i=1}^k \pi_i(\mathit{h}_l) \mathit{A}_i \mathit{B}_i^\top \mathit{h}_l + \bar{\mathit{R}} \mathit{h}_l$.
Distinct from LoRA parameters, which store specific semantic knowledge and are susceptible to error accumulation from continuous updating, the router learns a robust "Complexity-Capacity Mapping." Accordingly, we propose the CRU strategy to maintain the router's state throughout adaptation, empowering it to effectively capture fine-grained visual complexities for optimal rank selection.

\section{Theoretical Justification}
\label{sec:theory}

In this section, we provide a formal theoretical analysis to justify the optimization dynamics of MuRA framework. Specifically, we demonstrate why dynamic rank routing is mathematically necessary for diverse test streams and prove the stability of the CRU strategy introduced in Section~\ref{sec:method}.

To simplify our notation, let $f_i(\mathit{h}_l) = \mathit{A}_i \mathit{B}_i^\top \mathit{h}_l$ denote the isolated feature transformation of the $i$-th rank expert, and let $\pi_i = \boldsymbol{\pi}(\mathit{h}_l)_i$ denote its corresponding routing probability.

\begin{lemma}[Gradient Structure of the Router]
\label{lemma:gradient}
Applying the chain rule through the softmax routing mechanism, the gradient of the adaptation loss $\mathcal{L}$ with respect to the router weights $\mathit{W}_r^i$ for expert $i$ is formulated as:
$$\nabla_{\mathit{W}_r^i} \mathcal{L} = \mathbb{E} \left[ \pi_i(1 - \pi_i) \frac{\partial \mathcal{L}}{\partial \mathit{o}} f_i(\mathit{h}_l) \mathit{h}_l^\top - \sum_{j \neq i} \pi_j \pi_i \frac{\partial \mathcal{L}}{\partial \mathit{o}} f_j(\mathit{h}_l) \mathit{h}_l^\top \right]$$
\end{lemma}

\textit{Remark.} Lemma \ref{lemma:gradient} shows that the first term gives positive feedback when expert $i$ minimizes the loss, the second penalizes it by other experts' relative effectiveness, and jointly the router maximizes selecting the most effective expert for $\mathit{h}_l$.

To understand why a static rank is insufficient, we define $H(x)$ as the information entropy of the input $x$, which characterizes its complexity. We partition the data distribution into semantic regions $\Omega_i$ with concentrated entropy $H(\Omega_i)$ and associate each rank with a capacity center $\mu_i$ and local risk $R_i(H)=\alpha_i+\lambda_i\,d(H,\mu_i)^2$, where $d(H,\mu_i)$ measures complexity-capacity mismatch. Let $z_i=(\mathit{W}_r^i)^\top\mathit{h}_l$, $\pi_i=\exp(z_i)/\sum_j\exp(z_j)$, and $\bar f(\mathit{h}_l)=\sum_j\pi_j f_j(\mathit{h}_l)$. With $g=\partial\mathcal{L}/\partial\mathit{o}$, the chain rule gives $\partial\mathcal{L}/\partial z_i=\pi_i\langle g,f_i(\mathit{h}_l)-\bar f(\mathit{h}_l)\rangle$; using the first-order approximation $\langle g,f_i(\mathit{h}_l)-\bar f(\mathit{h}_l)\rangle\approx\ell_i(x)-\ell_\pi(x)$ connects the routing update to the hypothetical loss $\ell_i(x)$ of expert $i$ and the current mixture loss $\ell_\pi(x)$.

\begin{lemma}[The Necessity of Dynamic Routing]
\label{lemma:correlation}
One router gradient step yields
$$
\Delta z_i=-\eta_r\frac{\partial\mathcal{L}}{\partial z_i}\approx\eta_r\pi_i\big[\ell_\pi(x)-\ell_i(x)\big].
$$
Consequently, for a semantic region $\Omega_i$ with concentrated entropy $H(\Omega_i)$, the expected logit update satisfies
$$
\mathbb{E}_{x\in\Omega_i}[\Delta z_i]\propto\pi_i\Big[\sum_j\pi_j\,d(H(\Omega_i),\mu_j)^2-d(H(\Omega_i),\mu_i)^2\Big].
$$
\end{lemma}

\textit{Remark.} The update promotes an expert if and only if its complexity-capacity mismatch is below the routing-weighted average mismatch across all experts, thereby inducing explicit rank competition. Consequently, low-complexity inputs drift toward low-rank experts, whereas high-complexity inputs drift toward high-rank experts. This relative criterion replaces an isolated absolute-mismatch interpretation of entropy and capacity. A single static rank cannot participate in such competition and therefore forces suboptimal compromises across diverse inputs, leading to underfitting or overfitting. By decoupling representational capacity from individual image complexity, UCF enables token-specific capacity allocation.

Finally, we analyze the stability of maintaining the router's state across sequential test samples. Let $\mathit{W}_r^{(t)}$ denote the router weights at adaptation step $t$, $\eta_r$ denote the router's learning rate, and $L_\pi$ denote the Lipschitz constant of the softmax function.

\begin{lemma}[Stability of Continuous Router Updating]
\label{lemma:stability}
Given bounded gradients during test-time adaptation, the change in token-level routing probability across sequential updates is bounded by the Lipschitz continuity of the softmax distribution:
$$|\pi_i^{(t+1)}(\mathit{h}_l) - \pi_i^{(t)}(\mathit{h}_l)| \leq L_\pi \cdot \eta_r \cdot \| \nabla_{\mathit{W}_r} \mathcal{L} \|$$
\end{lemma}

\textit{Remark.} In standard joint-training MoE architectures, continuous sequential updates often cause routing collapse because the experts' representations constantly drift. However, as detailed in Section \ref{sec:method}, the MuRA experts ($\mathit{A}_i, \mathit{B}_i$) are explicitly reset to their MROD-initialized principal components after adapting to each sample. Because these capacity anchors remain fixed, the router's optimization landscape—the generalized "complexity-to-capacity" mapping described in Lemma \ref{lemma:correlation}—is globally stationary. Lemma \ref{lemma:stability} theoretically guarantees that accumulating gradients for the router over diverse test streams is mathematically stable. This proves that our Continuous Router Updating (CRU) strategy successfully aggregates structural routing knowledge without suffering from catastrophic forgetting or semantic drift.


\begin{table*}[t]
\centering
\caption{Performance comparison on ImageNet and its OOD variants.}
\label{tab:ood}
\renewcommand{\arraystretch}{1.0}
\resizebox{\textwidth}{!}{
\begin{tabular}{clcccccccc}
\toprule
Category & Method & ImageNet & ImageNet-A & ImageNet-V2 & ImageNet-R & ImageNet-S & Average & OOD Average \\
\midrule  
w/o TTA & CLIP-ViT-B/16\pub{ICML 22} & 68.34 & 49.89 & 61.88 & 77.65 & 48.24 & 61.20 & 59.42 \\
\midrule
\multirow{4}{*}{\makecell{Input \\ Adaptive}} 
& TPT\pub{NeurIPS 22} & 68.98 & 54.77 & 63.45 & 77.06 & 47.94 & 62.44 & 60.81 \\
& DiffTPT\pub{ICCV 23} & 70.30 & 55.68 & 65.10 & 75.00 & 46.80 & 62.28 & 60.52 \\
& PromptAlign\pub{NeurIPS 23} & - & 59.37 & 65.29 & 79.33 & 50.23 & - & 63.55 \\
& ADTE\pub{ICLR 26} & 71.80 & 65.50 & 65.60 & 81.40 & \textbf{53.50} & 67.50 & 66.50 \\
\midrule
\multirow{4}{*}{\makecell{Output \\ Adaptive}} 
& TDA\pub{CVPR 24} & 69.51 & 60.11 & 64.67 & 80.24 & 50.54 & 65.01 & 63.89 \\
& MTA\pub{CVPR 24} & 70.08 & 58.06 & 64.24 & 78.33 & 49.61 & 64.06 & 62.56 \\
& GS-Bias\pub{ICML 25} & 69.02 & 54.55 & 63.37 & 76.64 & 48.21 & 62.36 & 60.69 \\
& TT-RAA\pub{ICCV 25} & 70.23 & 60.59 & 64.69 & 80.58 & 49.98 & 65.01 & 63.96 \\
\midrule
\multirow{2}{*}{\makecell{Knowledge \\ Adaptive}} 
& TTL\pub{WACV 25} & 70.23 & 60.51 & 64.55 & 77.54 & 48.61 & 64.29 & 62.80 \\
& \textbf{MuRA (Ours)} & \textbf{72.46} & \textbf{66.15} & \textbf{66.48} & \textbf{82.47} & 51.76 & \textbf{67.86} & \textbf{66.72} \\
\bottomrule
\end{tabular}
}
\end{table*}

\begin{table*}[t]
\centering
\caption{Performance comparison on the Cross-Domain benchmark.}
\label{tab:cross-domain}
\renewcommand{\arraystretch}{1.3}
\resizebox{\textwidth}{!}{
\begin{tabular}{clccccccccccc}
\toprule
Category & Method & Aircraft & Caltech & Cars & DTD & EuroSAT & Flower & Food & Pets & SUN & UCF & Average \\
\midrule
w/o TTA & CLIP-ViT-B/16\pub{ICML 22} & 23.67 & 93.35 & 65.48 & 44.27 & 42.01 & 67.44 & 83.65 & 88.25 & 62.59 & 65.13 & 63.58 \\
\midrule
\multirow{4}{*}{\makecell{Input \\ Adaptive}} 
& TPT\pub{NeurIPS 22} & 24.78 & 94.16 & 66.87 & 47.75 & 42.44 & 68.98 & 84.67 & 87.79 & 65.50 & 68.04 & 65.10 \\
& DiffTPT\pub{ICCV 23} & 25.60 & 92.49 & 67.67 & 47.65 & 43.13 & 70.15 & \textbf{87.23} & 88.22 & 65.74 & 62.67 & 65.17 \\
& PromptAlign\pub{NeurIPS 23} & 24.80 & 94.01 & 68.50 & 47.24 & 47.86 & 72.39 & 86.65 & 90.76 & 67.54 & 69.47 & 66.92 \\
& ADTE\pub{ICLR 26} & 28.90 & 94.80 & \textbf{70.90} & 49.50 & 53.80 & 72.60 & 86.30 & 89.70 & \textbf{70.40} & \textbf{73.10} & 69.00 \\
\midrule
\multirow{4}{*}{\makecell{Output \\ Adaptive}} 
& TDA\pub{CVPR 24} & 23.91 & 94.24 & 67.28 & 47.40 & 58.00 & 71.42 & 86.14 & 88.63 & 67.62 & 70.68 & 67.53 \\
& MTA\pub{CVPR 24} & 25.20 & 94.21 & 68.47 & 45.90 & 45.36 & 68.06 & 85.00 & 88.24 & 66.67 & 68.69 & 65.58 \\
& GS-Bias\pub{ICML 25} & 25.30 & 94.16 & 66.77 & 45.10 & 43.63 & 68.86 & 85.67 & 88.58 & 64.78 & 65.74 & 64.86 \\
& TT-RAA\pub{ICCV 25} & 25.38 & 94.08 & 66.42 & 47.99 & \textbf{66.12} & 72.68 & 86.09 & 89.83 & 67.69 & 71.29 & 68.76 \\
\midrule
\multirow{2}{*}{\makecell{Knowledge \\ Adaptive}} 
& TTL\pub{WACV 25} & 23.82 & 93.63 & 67.97 & 46.69 & 42.02 & 70.48 & 85.05 & 88.72 & 66.32 & 69.20 & 65.39 \\
& \textbf{MuRA (Ours)} & \textbf{30.96} & \textbf{95.33} & 68.88 & \textbf{54.79} & 57.05 & \textbf{74.54} & 85.57 & \textbf{92.61} & 69.18 & 70.31 & \textbf{69.92} \\
\bottomrule
\end{tabular}
}
\end{table*}

\section{Experiments}
\subsection{Experimental Settings}

\paragraph{\textbf{Benchmarks.}} We evaluate MuRA on two distinct benchmarks: the out-of-distribution (OOD) benchmark and the cross-domain benchmark. The OOD benchmark assesses model robustness on four ImageNet~\citep{deng2009imagenet} variants: ImageNet-A~\citep{hendrycks2021natural}, ImageNet-V2~\citep{recht2019imagenet}, ImageNet-R~\citep{hendrycks2021many}, and ImageNet-S~\citep{wang2019learning}. The cross-domain benchmark evaluates adaptation capability across ten diverse domains: Aircraft~\citep{maji2013fine}, Caltech101~\citep{fei2004learning}, Cars~\citep{krause20133d}, DTD~\citep{cimpoi2014describing}, EuroSAT~\citep{helber2019eurosat}, Flower102~\citep{nilsback2008automated}, Food101~\citep{bossard2014food}, Pets~\citep{parkhi2012cats}, SUN397~\citep{xiao2010sun}, and UCF101~\citep{soomro2012ucf101}.

\paragraph{\textbf{Implementation Details.}} We build our framework upon the pre-trained CLIP model. The rank configuration is set to \{2,4,8,16,32\}. For each test image, we generate 63 augmented views. We optimize the parameters for a single step using the AdamW~\citep{loshchilov2017decoupled} optimizer, with learning rates of 6e-3 and 1e-4 for the low-rank adaptation matrices and the router, respectively.

\paragraph{\textbf{Compared Methods.}} We evaluate MuRA against the zero-shot CLIP baseline and three categories of state-of-the-art TTA approaches: (1) \textit{Input-adaptive} methods, including TPT, DiffTPT, PromptAlign, and ADTE~\citep{wu2026adaptive}; (2) \textit{Output-adaptive} methods, including TDA, MTA, GS-Bias~\citep{huang2025gs}, and TT-RAA~\citep{fan2025test}; and (3) \textit{Knowledge-adaptive} methods, specifically TTL alongside our MuRA.

\subsection{Main Results}

\paragraph{\textbf{Results on OOD Benchmark.}} As shown in Table~\ref{tab:ood}, MuRA achieves state-of-the-art performance across ImageNet and its out-of-distribution variants. This not only demonstrates vastly superior robustness compared to the static-rank approach TTL but also outperforms the recent strong baseline ADTE. While ADTE shows a slight edge on ImageNet-S, MuRA secures the highest overall average and OOD average, validating our core motivation that dynamic rank selection is essential for handling varying data complexities under distribution shifts.

\paragraph{\textbf{Results on Cross-Domain Benchmark.}} Table~\ref{tab:cross-domain} demonstrates MuRA's superior generalization capability across diverse visual domains. Compared to recent SOTA methods like ADTE and TT-RAA, which may excel in specific target datasets, MuRA exhibits a much more balanced and robust adaptation profile across different visual characteristics. Crucially, on datasets requiring fine-grained semantic extraction—such as the visually ambiguous Aircraft dataset and the texture-heavy DTD—MuRA's multi-rank mechanism yields striking improvements, substantially outperforming all competing methods. Ultimately, MuRA secures the highest average accuracy across all ten domains, further cementing the effectiveness of dynamic capacity allocation.

\subsection{Ablation Study}

In this section, we systematically ablate the core components of MuRA to understand their individual contributions. 

\paragraph{\textbf{Design Analysis.}} We first conduct comprehensive ablation studies across three challenging OOD datasets to validate the overall architecture. Table~\ref{tab:ablation} presents a systematic investigation of our main designs. The MROD-initialization brings significant improvements even in a single-rank setting, achieving a 7.38\% average accuracy gain over the vanilla CLIP. Furthermore, extending to multi-rank scenarios via UCF and incorporating the CRU strategy effectively leverages accumulated knowledge for better rank routing, leading to an additional 1.13\% improvement in average accuracy over the model with MROD and UCF only.

\begin{figure}[t] 
    \centering
    \renewcommand{\thesubfigure}{\alph{subfigure}}
    
    \begin{subfigure}[b]{0.235\textwidth} 
        \centering
        \includegraphics[width=\textwidth]{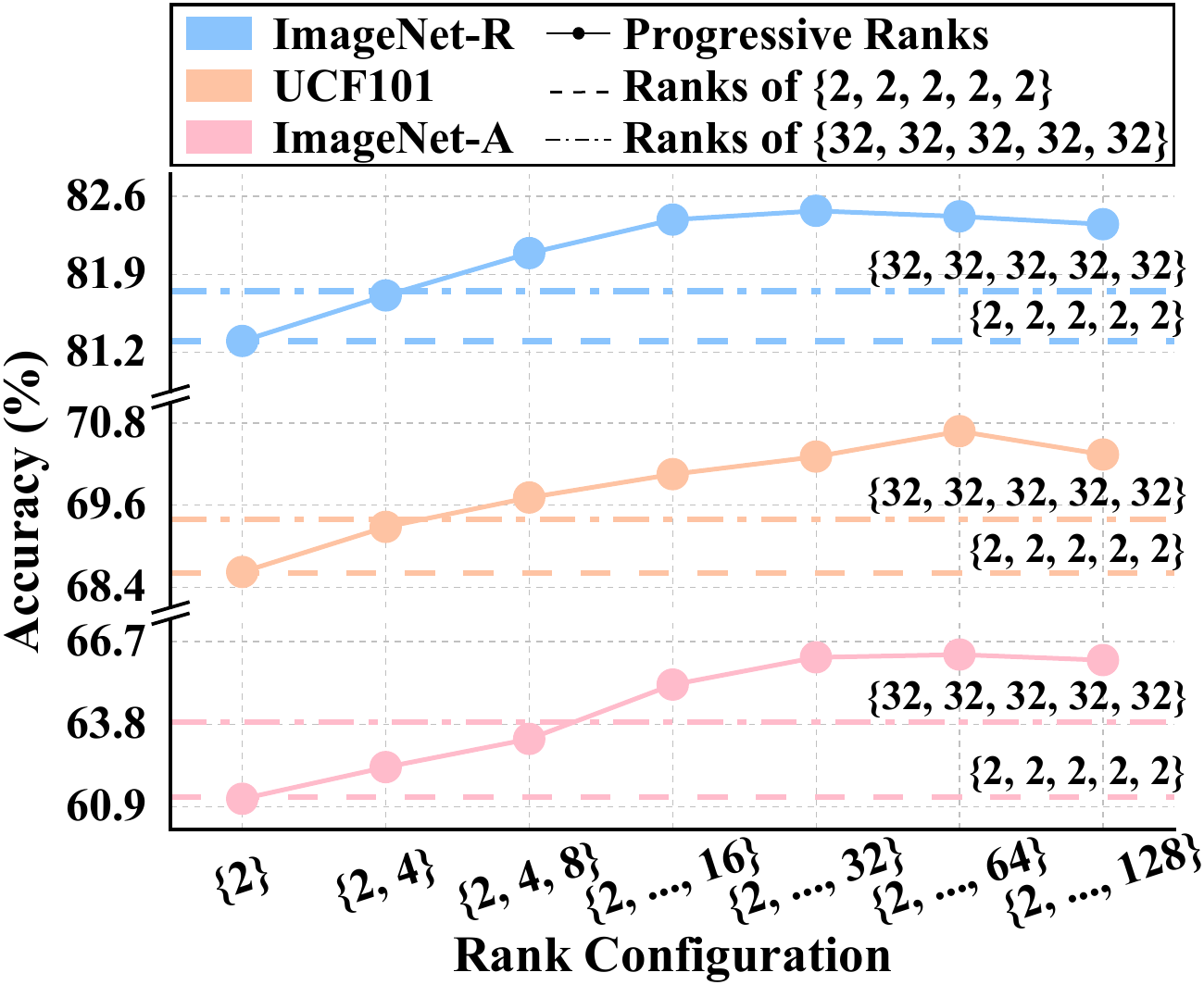}
        \caption{Rank configuration analysis.}
        \label{fig:ranks}
    \end{subfigure}
    \hfill
    \begin{subfigure}[b]{0.235\textwidth}
        \centering
        \includegraphics[width=\textwidth]{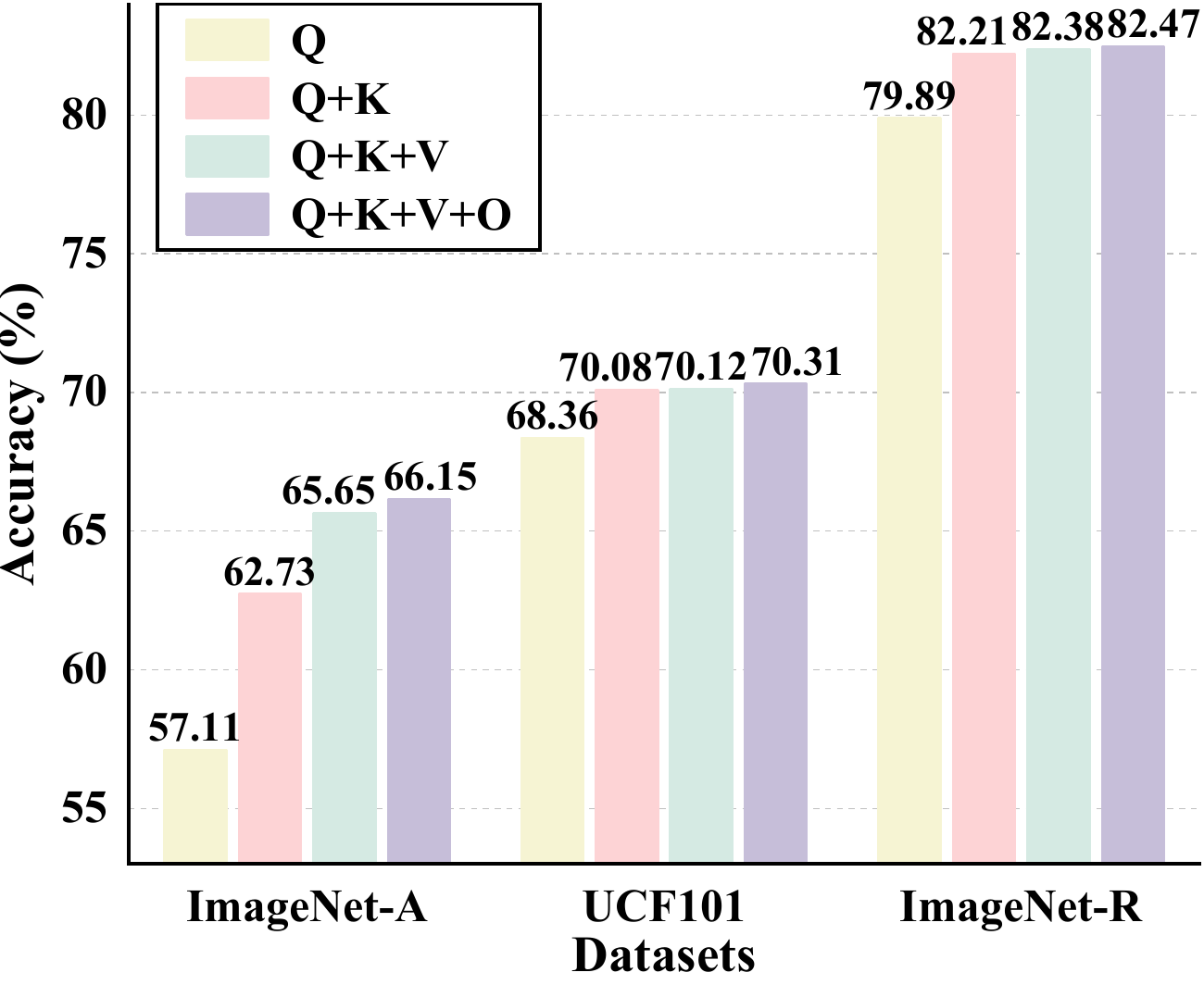}
        \caption{Attention weights to adapt.}
        \label{fig:params}
    \end{subfigure}
    
    \caption{Analysis of (a) different rank configurations, where solid lines represent rank configuration while dashed lines indicate homogeneous low-rank \{2,2,2,2,2\} and high-rank \{32,32,32,32,32\} settings, (b) The impact of adapting different attention matrices.}
    \label{fig:analysis}
\end{figure}

\paragraph{\textbf{Rank Configurations and Attention Adaptation.}} Having established the effectiveness of the overall framework, we investigate the specific configurations of the low-rank modules. In Figure~\ref{fig:ranks}, performance curves show significant improvements as ranks progressively increase up to \{2,4,8,16,32\}, beyond which the gains plateau while computational costs continue to rise. The substantial gap between our progressive strategy and homogeneous configurations validates that adapting to varying data complexities requires a spectrum of rank components. Furthermore, as shown in Figure~\ref{fig:params}, progressively incorporating more attention matrices—from solely the query matrix (Q) to the full set (Q, K, V, O)—yields consistent performance improvements, confirming our strategy of updating the complete attention block to maximize adaptation efficacy.

\begin{table}[t] 
\centering
\small 
\caption{Design ablation study of our MuRA.} 
\label{tab:ablation}
\setlength{\tabcolsep}{6pt} 
\renewcommand{\arraystretch}{1.0}
\begin{tabular}{ccc|cccc}
\toprule  
\multicolumn{3}{c|}{Main Designs} & \multirow{2}{*}{Img-A} & \multirow{2}{*}{Img-R} & \multirow{2}{*}{UCF} & \multirow{2}{*}{Avg.} \\
\cline{1-3}
MROD & UCF & CRU & & & & \\
\midrule 
$\times$ & $\times$ & $\times$ & 49.89 & 77.65 & 65.13 & 64.22 \\
$\times$ & $\checkmark$ & $\times$ & 60.61 & 81.26 & 69.15 & 70.34 \\
$\checkmark$ & $\times$ & $\times$ & 63.93 & 81.63 & 69.23  & 71.60 \\
$\checkmark$ & $\checkmark$ & $\times$ & 64.24 & 81.85 & 69.47 & 71.85 \\
$\checkmark$ & $\checkmark$ & $\checkmark$ & \textbf{66.15} & \textbf{82.47} & \textbf{70.31} & \textbf{72.98} \\
\bottomrule
\end{tabular}
\end{table}

\begin{table}[t]
\centering
\caption{Performance comparison of Single-Rank variants vs. MuRA across different depths (Average accuracy on ImageNet-A, ImageNet-R, and UCF101). Layer groups: Bottom (1-4), Mid (5-8), Neck (9-10), Penultimate (11), and Deepest (12).}
\label{tab:layer_depth_comparison}
\resizebox{\linewidth}{!}{
\begin{tabular}{l|ccccc}
\toprule
Method & Bottom & Mid & Neck & Penultimate & Deepest \\
\midrule
Single-Rank & \textbf{67.87} & 67.46 & 67.71 & 67.78 & 67.51 \\
Single-Rank (SVD-Init) & 60.33 & 55.23 & \textbf{70.74} & 70.46 & 70.30 \\
\textbf{MuRA (Ours)} & 60.08 & 56.76 & 70.56 & 70.77 & \textbf{72.98} \\
\bottomrule
\end{tabular}
}
\end{table}

\begin{table}[t]
    \centering
    \small
    \renewcommand{\arraystretch}{1.2} 
    \caption{Ablation of different routing strategies.}
    \label{tab:routing_comparison}
    \begin{tabular}{l|cccc}
        \toprule
        Routing Strategy & ImageNet-A & ImageNet-R & UCF101 & Avg. \\
        \midrule
        Hard Routing & 63.17 & 81.66 & 69.89 & 71.57 \\
        \textbf{Soft Routing} & \textbf{66.15} & \textbf{82.47} & \textbf{70.31} & \textbf{72.98} \\
        \bottomrule
    \end{tabular}
\end{table}

\begin{table}[t]
    \centering
    \small
    \renewcommand{\arraystretch}{1.2} 
    \caption{Ablation study of different router designs. mAcc represents the mean accuracy across ImageNet-A, ImageNet-R, and UCF101 datasets.}
    \label{tab:router}
    \begin{tabular}{l|ccc}
        \toprule
        \multirow{2}{*}{Router Design} & \multicolumn{3}{c}{mAcc} \\
        \cline{2-4}
        & w/o CRU & \textbf{w CRU} & $\Delta$ (\%) \\
        \midrule
        Instance-Level & 71.86 & 71.89 & +0.03 \\
        \textbf{Token-Level} & 71.85 & \textbf{72.98} & \textbf{+1.13} \\
        \bottomrule
    \end{tabular}
\end{table}

\paragraph{\textbf{Routing Mechanism Strategies.}} We further examine the design of the Unified Component Fusion (UCF) router by evaluating its operational granularity and routing strategy. First, regarding granularity, Table~\ref{tab:router} compares Instance-Level and Token-Level routers. When CRU is enabled, the Token-Level router shows significant gains (+1.13\% mAcc), demonstrating that effective rank adaptation requires token-level granularity to capture local variations in visual complexity. Second, regarding the routing strategy, Table~\ref{tab:routing_comparison} shows that Soft Routing consistently outperforms Hard Routing across all datasets, validating that combining multiple rank components flexibly is more effective than strictly selecting a single rank.

\subsection{Efficiency and Scalability Analysis}
A practical test-time adaptation method must not only be accurate but also structurally efficient and scalable to larger foundation models. In this section, we analyze the source of MuRA's efficiency, its computational footprint, and its scalability.

\paragraph{\textbf{Adaptation Depth and Inherent Efficiency.}} 
Table~\ref{tab:layer_depth_comparison} shows that standard Single-Rank drops from 67.87\% at Bottom to 67.51\% at Deepest, whereas MuRA's dynamic rank routing reverses this trend and peaks at Deepest; parameter-matched SVD-initialized Single-Rank improves deeper adaptation over standard-LoRA Single-Rank but peaks at Neck. Static-rank recovery thus requires shallower adaptation, forfeiting the fastest updates and lowest memory from the shortest backpropagation path, supporting dynamic ranks as the source of MuRA's effective, efficient deepest-layer adaptation.

\begin{table}[t]
\centering
\caption{Efficiency and performance comparison. Throughput is measured in samples/second. CLIP$^\dagger$ denotes the zero-shot baseline (theoretical upper bound for efficiency). mAcc is the mean accuracy across ImageNet-A, -R, and UCF101. Best adaptation results are in \textbf{bold}.}
\label{tab:efficiency_detailed}
\resizebox{\columnwidth}{!}{
\begin{tabular}{lcccc}
\toprule
Method & FLOPs (T) & Memory (GB) & Throughput & mAcc \\
\midrule
CLIP$^\dagger$ & 2.25 & 0.67 & 19.57 & 64.22 \\
\midrule
TPT & 3.32 & 4.34 & 3.20 & 66.62 \\
Single-Rank (Matched) & \textbf{2.25} & 3.08 & 4.59 & 68.21 \\
\textbf{MuRA (Ours)} & 2.40 & \textbf{2.05} & \textbf{11.26} & \textbf{72.98} \\
\bottomrule
\end{tabular}
}
\end{table}

\begin{table}[t]
\centering
\small
\caption{Performance comparison using the larger ViT-L/14 backbone. The best results are highlighted in \textbf{bold}.}
\label{tab:large_model}
\resizebox{\linewidth}{!}{
\begin{tabular}{l|ccccccc}
\toprule
Method & ImageNet & Img-A & Img-V2 & Img-R & Img-S & Avg & OOD Avg. \\
\midrule
CLIP-ViT-L/14 & 74.04 & 53.88 & 67.69 & 87.42 & 63.18 & 69.31 & 68.13 \\
Single-Rank & 75.62 & 65.21 & 69.43 & 87.46 & 63.51 & 72.25 & 71.40 \\
\textbf{MuRA (Ours)} & \textbf{78.66} & \textbf{81.96} & \textbf{73.19} & \textbf{91.79} & \textbf{64.69} & \textbf{78.06} & \textbf{77.91} \\
\bottomrule
\end{tabular}
}
\end{table}

\paragraph{\textbf{Efficiency-Effectiveness Trade-off.}} Building upon its deep-layer design, MuRA unlocks an optimal trade-off. As detailed in Table~\ref{tab:efficiency_detailed}, while conventional methods like TPT suffer from severe throughput bottlenecks and memory bloat, MuRA maintains a highly practical throughput and requires only 2.05 GB of memory—closely mirroring the efficiency profile of zero-shot CLIP. At this minimal computational cost, MuRA yields a massive accuracy improvement over the baseline. Crucially, we evaluate MuRA against a \textit{Single-Rank (Matched)} baseline configured with an identical number of trainable parameters. Against a \textit{Single-Rank (Matched)} baseline with an identical number of trainable parameters, MuRA delivers a 4.77\% accuracy gain, runs more than twice as fast, and uses significantly less memory, confirming that gains stem from dynamic routing rather than naive capacity increase.

\begin{table}[t]
    \centering
    \small
    \caption{Performance comparison on the CLIP-ResNet-50 backbone. Avg. is the mean accuracy over ImageNet-A, ImageNet-R, and UCF101. Best results are in \textbf{bold}.}
    \label{tab:resnet50}
    \resizebox{\linewidth}{!}{
        \begin{tabular}{l|cccc}
            \toprule
            Method & ImageNet-A & ImageNet-R & UCF101 & Avg. \\
            \midrule
            CLIP & 21.83 & 56.15 & 58.84 & 45.61 \\
            \midrule
            TPT & 26.67 & 59.11 & 60.82 & 48.87 \\
            TDA & 30.29 & 62.58 & 64.18 & 52.35 \\
            Single-Rank & 27.41 & 60.37 & 59.95 & 49.24 \\
            Single-Rank (SVD-Init) & 29.18 & 61.26 & 62.25 & 50.90 \\
            MuRA (Instance-Level) & 30.55 & 62.79 & 64.30 & 52.55 \\
            \textbf{MuRA (Token-Level)} & \textbf{32.86} & \textbf{63.14} & \textbf{65.76} & \textbf{53.92} \\
            \bottomrule
        \end{tabular}
    }
\end{table}

\paragraph{\textbf{Scalability to Larger Models.}} 
Evaluating on the larger ViT-L/14 backbone (Table~\ref{tab:large_model}), MuRA achieves a superior 78.06\% average accuracy. The performance leap is most striking on ImageNet-A, where MuRA reaches 81.96\%—yielding a massive +28.08\% absolute gain over vanilla CLIP and outperforming the static Single-Rank baseline by 16.75\%. These results demonstrate that while fixed-rank constraints bottleneck larger architectures, MuRA effectively unleashes their full representational potential.

\begin{figure}[t]
    \centering
    \includegraphics[width=1\linewidth]{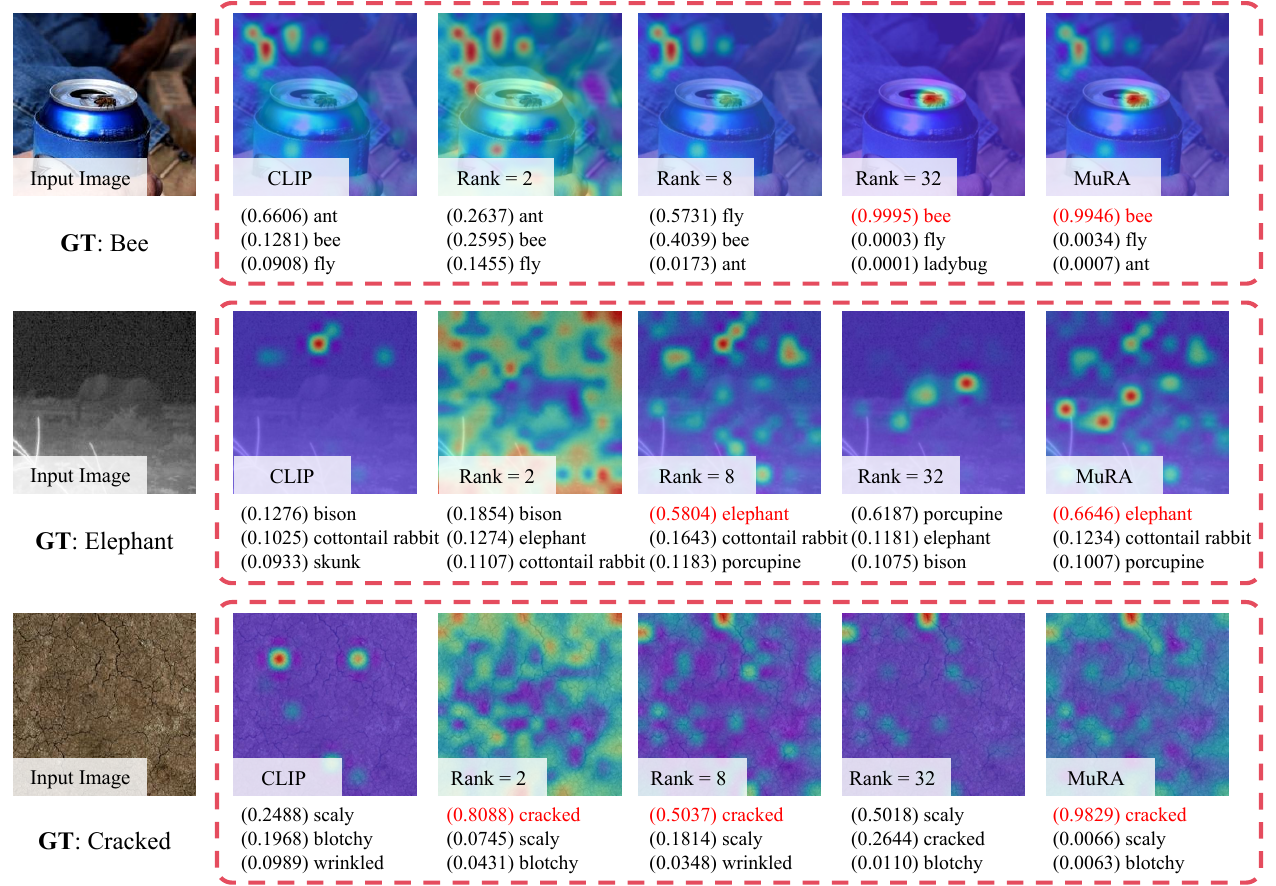}
    \caption{Visualization of attention maps and top-3 predictions for CLIP, different rank components, and MuRA.}
    \label{fig:attn_map}
\end{figure}

\paragraph{\textbf{Generalization to CNN Backbones.}}
We evaluate CLIP-ResNet-50 under the same protocol (Table~\ref{tab:resnet50}). MuRA adapts final-bottleneck $1{\times}1$ convolutions. Single-Rank uses rank 64 on these layers, matching 0.33M. Token-level routing treats each spatial location's channel-wise vector as a token. MuRA (Token-Level) achieves the best mean accuracy, surpassing TPT, TDA, both single-rank baselines, and instance-level routing, demonstrating that MuRA generalizes across backbones rather than being specific to ViTs.


\begin{figure}[t]
    \centering
    \begin{subfigure}[b]{0.32\linewidth}
        \centering
        \includegraphics[width=\linewidth, trim=18 10 10 18, clip]{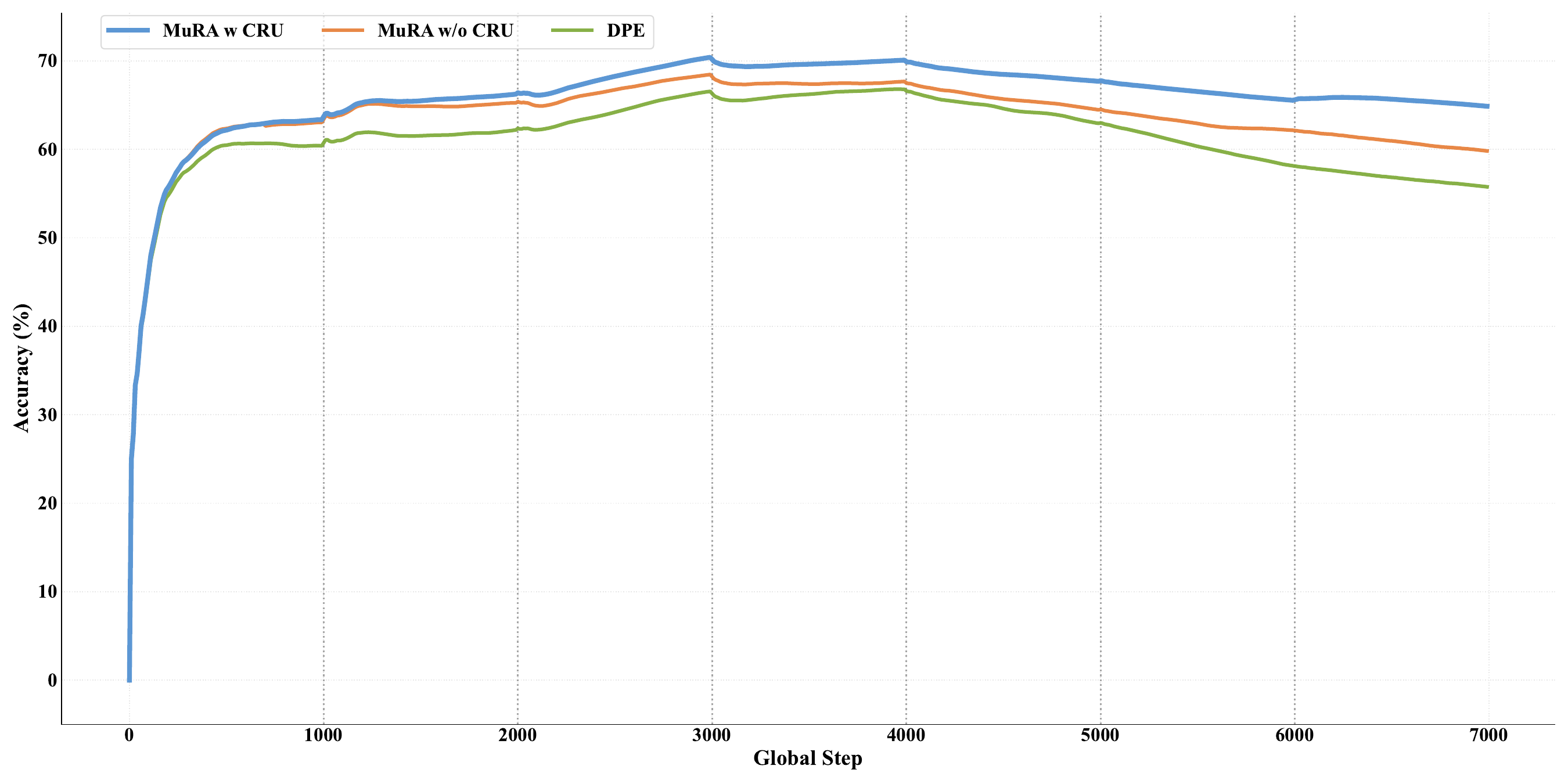}
        \caption{Accuracy.}
        \label{fig:cl_acc}
    \end{subfigure}\hfill
    \begin{subfigure}[b]{0.32\linewidth}
        \centering
        \includegraphics[width=\linewidth, trim=18 10 10 18, clip]{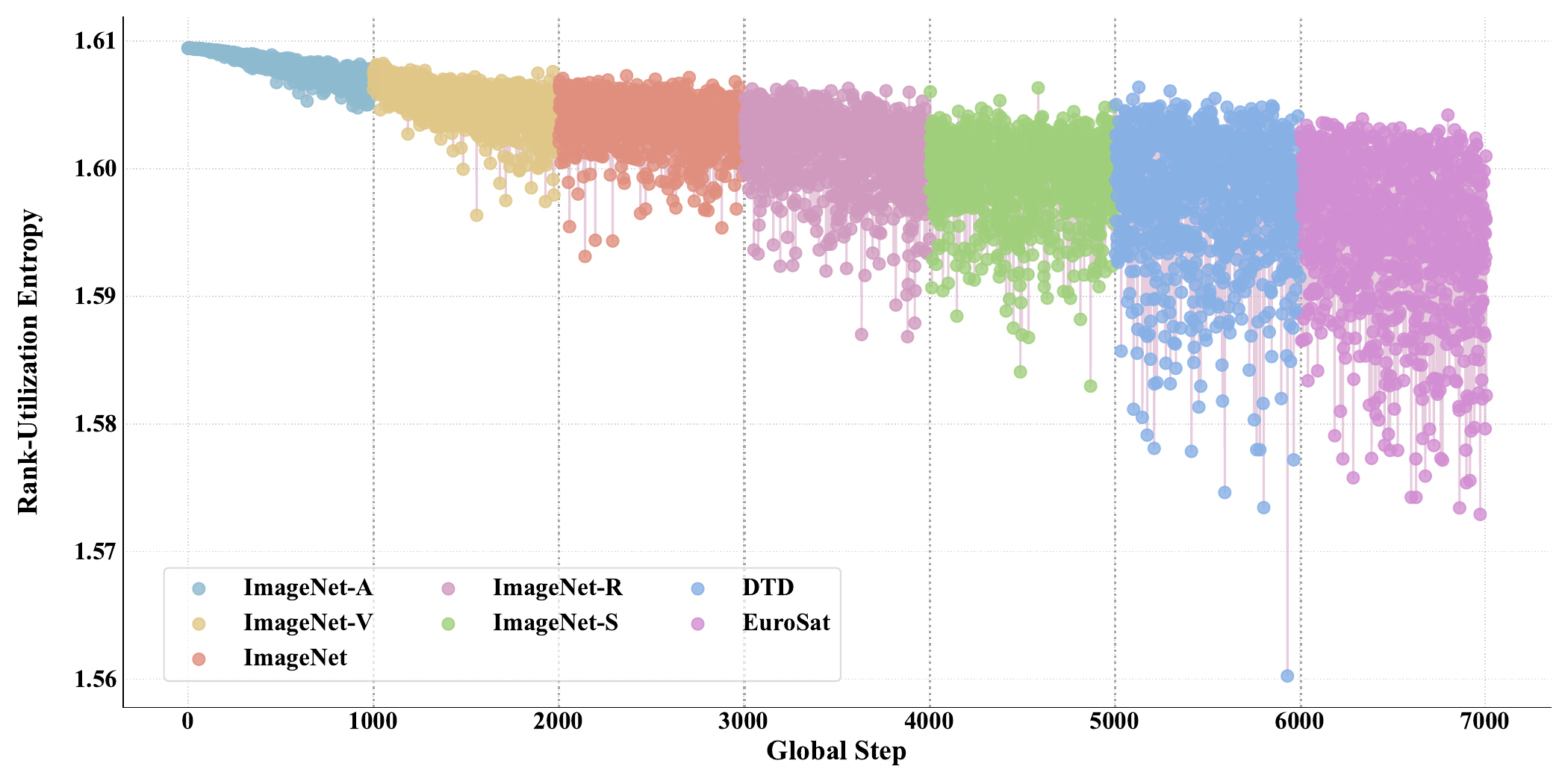}
        \caption{Rank entropy.}
        \label{fig:cl_entropy_main}
    \end{subfigure}\hfill
    \begin{subfigure}[b]{0.32\linewidth}
        \centering
        \includegraphics[width=\linewidth, trim=24 10 10 18, clip]{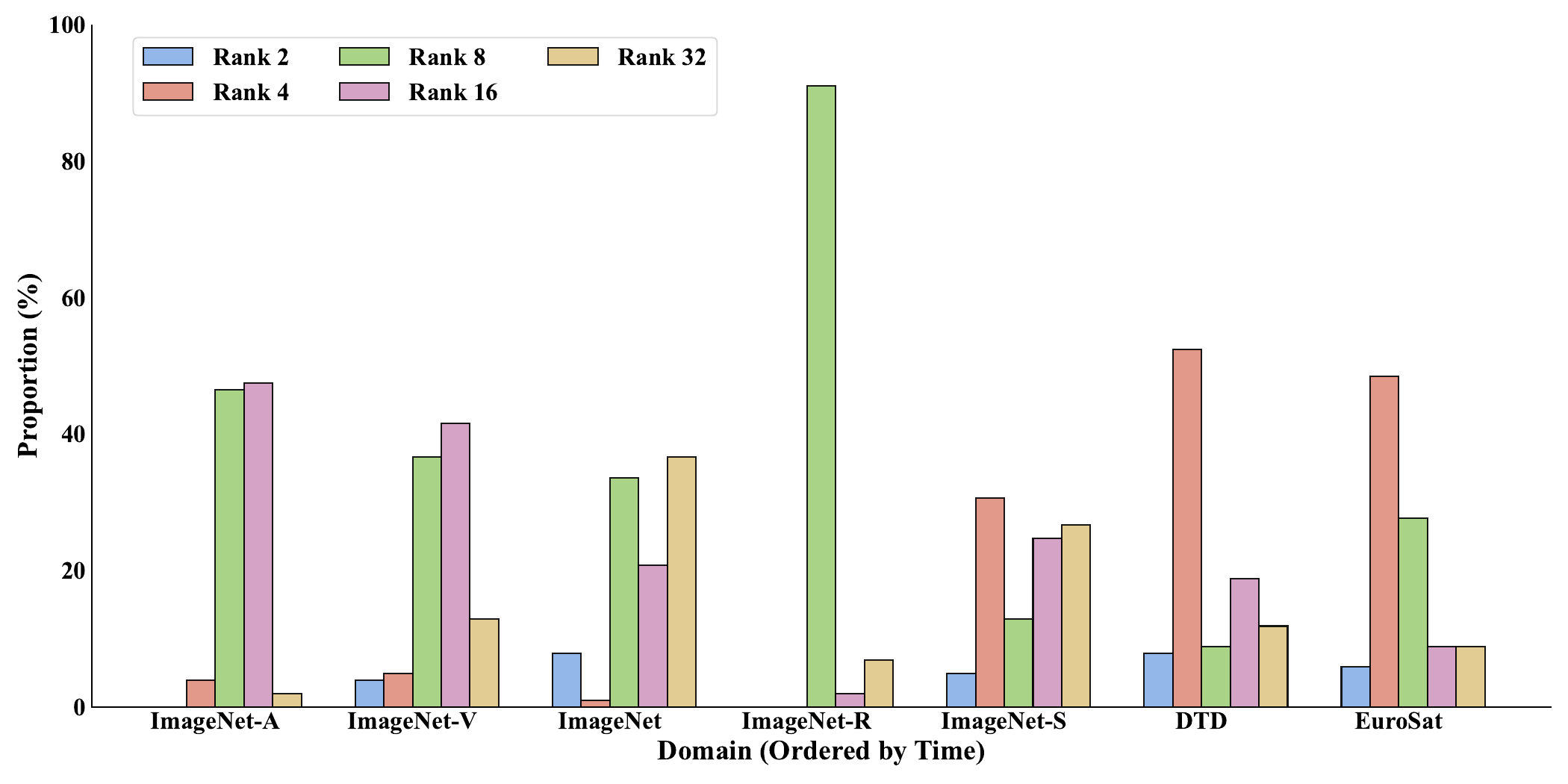}
        \caption{Routing profile.}
        \label{fig:cl_profile}
    \end{subfigure}
    \caption{\textbf{MuRA dynamically stabilizes and adapts its rank allocation under continuous distribution shifts.} Accuracy and Q-matrix routing dynamics are shown throughout the seven-domain stream.}
    \label{fig:cl_dynamics}
\end{figure}

\subsection{Qualitative Results}

\paragraph{\textbf{Attention Analysis.}} To qualitatively assess MuRA, we visualize the self-attention heatmaps of CLIP, MuRA, and the individual rank components within MuRA, along with their top-3 prediction probabilities. As shown in Figure~\ref{fig:attn_map}, different rank-specific LoRA components focus on distinct types of image information: low-rank matrices tend to produce diffuse attention over background or coarse features, while high-rank ones concentrate on semantically rich foreground areas. Importantly, MuRA dynamically selects appropriate rank components. For object-centric samples like the bee (Row 1), it leverages higher-rank components (Rank-32) for semantic-rich foreground attention. In contrast, for texture-centric samples like the cracked earth (Row 3), it utilizes lower-rank components (Rank-2) to capture broader attention distributions. CLIP's attention appears confined, limiting the effectiveness of purely output-adaptive methods.



\subsection{Router Dynamics under Continuous Distribution Shifts}

We evaluate MuRA on a continual TTA stream of seven sequential domains in decreasing intrinsic complexity---ImageNet-A, ImageNet-V, ImageNet, ImageNet-R, ImageNet-S, DTD, and EuroSAT---with 1,000 steps per domain, yielding a highly non-i.i.d., class-imbalanced stream with abrupt shifts. Figure~\ref{fig:cl_acc} shows MuRA with CRU achieving the highest accuracy throughout, surpassing MuRA without CRU and DPE. With CRU, rank-utilization entropy decreases (Figure~\ref{fig:cl_entropy_main}), indicating router stabilization and greater rank-selection certainty without cold start. Figure~\ref{fig:cl_profile} shows complex domains selecting ranks 8, 16, 32 and simpler domains ranks 2, 4; full Q/K/V/O results appear in the supplementary material.

\section{Conclusion and Future Work}

MuRA enables efficient and effective TTA by tailoring representational capacity to token-level input complexity; together, MROD, UCF, and CRU achieve state-of-the-art generalization at minimal cost. However, MuRA still uses manually specified rank boundaries, relies on a minimum-entropy objective that can induce over-confident predictions, and has not yet been validated on autoregressive LVLMs. Future work will derive boundaries from MROD singular-value decay while preserving pretrained energy; extend complexity-aligned capacity to autoregressive LVLM decoding for robust, hallucination-free reasoning; and develop optimization-driven gradient and capacity modulation to address over-confidence and generalize MuRA to broader architectures.

\begin{acks}
This work was supported in part by the National Natural Science Foundation of China under Grant U24B6012, 62406167.
\end{acks}



\bibliographystyle{ACM-Reference-Format}
\balance
\bibliography{sample-base}










\end{document}